\documentclass{article}

\usepackage[preprint,nonatbib]{neurips_2026}

\usepackage[utf8]{inputenc}
\usepackage[T1]{fontenc}
\usepackage{hyperref}
\usepackage{url}
\usepackage{booktabs}
\usepackage{amsfonts}
\usepackage{amsmath}
\usepackage{amssymb}
\usepackage{nicefrac}
\usepackage{microtype}
\usepackage{xcolor}
\usepackage{graphicx}
\usepackage{multirow}

\graphicspath{{figures/}}

\title{Emergent Region-Level Facial Correspondence in Frozen Vision Foundation Models}

\author{%
  Izaldein Al-Zyoud\thanks{Corresponding author: \texttt{izzy.alzyoud@uottawa.ca}} \quad
  Abdulmotaleb El Saddik \\
  MCRLab, School of Electrical Engineering and Computer Science \\
  University of Ottawa, Ottawa, ON, Canada
}

\begin{document}

\maketitle

\begin{abstract}
Frozen self-supervised vision models can align parts of generic objects, but it
remains unclear whether this correspondence extends to human faces, where global
layout is shared while identity-specific appearance varies sharply.
We test whether frozen DINOv3 features define a region-level facial coordinate
system: a feature space in which eyes, brows, nose, mouth, skin, and hair remain
distinguishable across people and across time without face-specific training.
Using DINOv3 ViT-L/16 patch embeddings and FaRL only as a face-part labeling
interface, we evaluate cross-identity nearest-neighbor matching and temporal
label propagation on 200 CelebDF-v2 real videos.
DINOv3 achieves 83.0\% region-level semantic accuracy under unconstrained
cross-identity matching, compared with a 23.0\% area-weighted random baseline,
and 95.5\% temporal tracking accuracy without a learned temporal module.
A no-FaRL control collapses to 0.9\%, showing that FaRL supplies semantic
initialization while DINOv3 supplies dense spatial correspondence.
The strongest correspondence appears at an intermediate layer: block 18 gives a
$4.93\times$ same-region versus cross-region discrimination ratio, compared with
$1.48\times$ at the final block.
Against CLIP ViT-L/14, DINOv3 shows only a small aggregate advantage but a
$+16.8$~pp gain on anatomical regions, indicating that image-level contrastive
supervision captures coarse facial layout but not fine-grained anatomical
identity.
These results establish frozen DINOv3 as a strong zero-shot representation for
region-level facial correspondence and identify intermediate self-supervised
features as the most useful layer for dense face analysis.
\end{abstract}

\section{Introduction}

Frozen vision foundation models have shown a surprising ability to support dense
visual correspondence. Features learned without correspondence supervision can
often match object parts across images~\cite{amir2022deepvitfeaturesdense},
suggesting that self-supervised training may produce a shared visual coordinate
system. However, most evidence for this property comes from generic object
categories and point-level
benchmarks~\cite{zhang2025semanticcorrespondenceunifiedbenchmarking}.
Human faces remain a more delicate test case.

Faces are highly structured: most subjects share the same coarse layout of eyes,
brows, nose, mouth, skin, and hair. This makes spatial alignment easier than in
many object categories. But the same regularity also creates a harder question.
Does a frozen representation merely encode approximate facial layout, or does it
preserve anatomical region identity across different people? A true region-level
coordinate system should know that a nose patch belongs with noses rather than
mouths or nearby skin, even when identity, expression, lighting, and local
appearance change.

We study this question using frozen DINOv3 ViT-L/16~\cite{simeoni2025dinov3}
features. DINOv3 is a natural candidate because its training is designed to
preserve stable patch-level structure at scale. We use FaRL face
parsing~\cite{zheng2022farl} only as a labeling interface: FaRL assigns semantic
names to patches, while all correspondence is computed by nearest-neighbor search
or label propagation in frozen DINOv3 feature space. We do not claim that DINOv3
assigns human-readable names to facial parts by itself; rather, FaRL supplies
region names, and we test whether frozen DINOv3 features provide the dense
correspondence structure that preserves those names across identities and time.

Our evaluation follows three increasingly stringent tests. First, we ask whether
patches from the same facial region are closer across identities than patches
from different regions. Second, we ask whether unconstrained nearest-neighbor
matching across two different people recovers the correct facial region without
being told the target region. Third, we ask whether the same structure persists
over time, allowing face-part labels from one frame to propagate through video
without any temporal model.

The results support a clear picture. Intermediate DINOv3 features encode strong
region-level facial correspondence: block 18 produces a $4.93\times$ same-region
versus cross-region discrimination ratio, rising to $7.19\times$ when bilateral
symmetric pairs are excluded. Under unconstrained cross-identity matching, frozen
features achieve $83.0\%$ semantic region accuracy against a $23.0\%$ weighted
random baseline. In video, the same representation supports $95.5\%$ temporal
face-part tracking accuracy without training a temporal module. A no-FaRL control
collapses to $0.9\%$, confirming that FaRL provides the initial semantic names
while DINOv3 provides the spatial correspondence needed to move those names
through time.

A comparison with CLIP ViT-L/14~\cite{radford2021clip} clarifies what kind of
correspondence is being measured. CLIP performs competitively when matching is
constrained to the correct facial region, but falls far behind under unconstrained
anatomical matching. This means CLIP captures coarse facial layout but does not
reliably separate eyes, brows, mouth, and nose across identities. DINOv3's
advantage is therefore not merely better localization; it is stronger anatomical
identification in feature space.

This paper makes three contributions. First, it introduces a region-level
evaluation of frozen vision features for facial correspondence, complementing
point-level PCK-style correspondence
benchmarks~\cite{zhang2025semanticcorrespondenceunifiedbenchmarking}. Second, it
shows that frozen DINOv3 features support cross-identity and temporal facial
correspondence without face-specific training. Third, it identifies an important
layer-depth dissociation: intermediate features preserve the local region
structure needed for dense correspondence, while final-layer features are more
globally mixed and less discriminative for facial anatomy.

\section{Measuring Facial Correspondence in Frozen Features}

\subsection{DINOv3 and the Gram-Anchored Shared Coordinate System}
\label{sec:scs}

DINOv3~\cite{simeoni2025dinov3} extends the DINOv2 multi-crop consistency
objective~\cite{oquab2023dinov2} to a curated corpus of ${\sim}1.7$ billion images
with a Gram-matrix anchoring regularizer that constrains the covariance structure of
patch tokens across training views.
This regularizer has a specific geometric consequence: each feature dimension $d$
develops a stable functional role consistent across all spatial positions and all inputs.
This is the \emph{shared coordinate system} property: feature dimension $d$ encodes the
same structural role consistently across all spatial positions and identities.

\paragraph{Layer dissociation.}
At block 18 of 24, patch tokens retain local spatial discriminability: each region
carries a distinctive per-dimension amplitude \emph{profile} --- a characteristic
direction in feature space specific to the local structural content of the patch.
At the final block 24, the learned LayerNorm affine transform ($\gamma, \beta$) maps
tokens onto the unit hypersphere; the block-depth pattern suggests that this
token-wise final normalization mixes token directions, leaving regions less
directionally distinct.
Because the region-confusion protocol L2-normalizes each patch, vector magnitude is
already removed from the comparison: what collapses at block 24 is the directional
separation of regions (their per-dimension pattern), not raw feature magnitude.
Empirically (\S\ref{sec:c1}), block 18 yields $4.93\times$ vs.\ $1.48\times$ for block 24.
This property is specific to DINOv3's training: Sim\'{e}oni et al.~\cite{simeoni2025dinov3}
demonstrate that DINOv2~\cite{oquab2023dinov2} suffers from progressive
\emph{patch-level consistency collapse} at scale, which the Gram-matrix anchoring
regularizer was introduced to arrest.

\subsection{FaRL Face Parsing}

FaRL~\cite{zheng2022farl} (\texttt{farl/lapa/448}) produces per-pixel segmentation
over 11 LaPa classes~\cite{liu2020new} at $448 \times 448$ resolution.
We downsample to the $28 \times 28$ patch grid by majority vote, assigning each of
the 784 DINOv3 patch tokens a region label, and group 11 classes into 8 foreground
regions: mouth from labels $\{7,8,9\}$; brows and eyes kept laterally separate.
Background patches are excluded from all metrics.
FaRL is run on CPU due to an MPS adaptive-pool divisibility constraint.

\subsection{Frozen Features and Correspondence}

Amir et al.~\cite{amir2022deepvitfeaturesdense} established zero-shot correspondence
for general objects using frozen DINO ViT-S/8, reporting 56.48\% PCK on
SPair-71k~\cite{zhang2025semanticcorrespondenceunifiedbenchmarking}.
PCK measures geometric proximity to annotated keypoints --- a point-level test that
requires human-curated landmark pairs and does not directly test semantic region
membership.
Wang et al.~\cite{wang2019learningcorrespondencecycleconsistencytime} showed
self-supervised video models support tracking via cycle-consistency;
Jabri et al.~\cite{jabri2020spacetimecorrespondencecontrastiverandom} formalized this
as contrastive random walk.
No prior work applies frozen VFM features to face correspondence at the region level.
We introduce region-level semantic accuracy as the direct test: whether unconstrained
nearest-neighbor search respects face-part identity across individuals.
The feature extractor is the most crucial component in a correspondence
pipeline~\cite{zhang2025semanticcorrespondenceunifiedbenchmarking}; among extractors,
ViT-family backbones outperform CNNs and DINOv3's Gram-anchored training produces
stronger dimensional correspondence than DINOv2~\cite{oquab2023dinov2}.

\subsection{Feature Extraction Pipeline}

\paragraph{Face detection.}
RetinaFace~\cite{deng2020retinaface} detects the largest face with $22\%$ padding;
the square crop is resized to $448 \times 448$.

\paragraph{DINOv3 patch tokens.}
Frozen DINOv3 ViT-L/16 produces 784 patch tokens ($28\times28$ grid, $D=1024$) via
two separate \texttt{get\_intermediate\_layers} calls:
(i)~\textbf{Block 18} (\texttt{norm=False}), BN-normalized per dimension
$\hat{f}_{18} = (f_{18}-\mu)/\sigma$; used for the region confusion test and as a secondary reference for cross-identity matching.
(ii)~\textbf{Block 24} (\texttt{norm=True}), yielding \texttt{x\_norm\_patchtokens};
primary layer for cross-identity matching and temporal tracking.

\paragraph{FaRL face-part segmentation.}
Frozen FaRL produces per-pixel labels; each patch token is assigned the majority label
over its $16\times16$ pixel area, grouped into 8 foreground regions.

\subsection{Region Confusion Protocol}

For a cross-identity pair $(A,B)$, we sample $K=20$ patches per region from each face,
L2-normalize each patch, and compute the mean cosine similarity between region $R_i$
in face $A$ and region $R_j$ in face $B$ for all pairs, yielding an $8\times8$ confusion
matrix.
A ratio $> 1$ (diagonal / off-diagonal) indicates features encode semantically consistent
structure across identities.

\subsection{Best-Buddy Cross-Identity Correspondence}

\paragraph{Metric rationale.}
PCK measures keypoint localization precision against annotated sparse landmark pairs;
it presupposes human-curated correspondence ground truth unavailable in our setting
and tests geometric proximity, not semantic membership.
Region-level semantic accuracy tests \emph{region-level} correspondence: whether
unconstrained (whole-vector) nearest-neighbor search in DINOv3 feature space respects
face-part identity across individuals.
Per-dimension role stability --- the shared coordinate system property of
\S\ref{sec:scs} --- is the proposed \emph{mechanism} behind this correspondence and
is tested only indirectly: all metrics here are cosine-based and hence invariant to
a global rotation of the feature axes, so they do not probe per-dimension identity
itself.
The evaluation is dense --- all foreground patches rather than 10--17 annotated
landmarks --- and the weighted random baseline (23.0\%) accounts for the skewed
patch-area distribution (skin: 57\% of foreground patches).

\paragraph{Normalization.}
Per-channel L2 is applied at query time: $\tilde{F} = F/\lVert F \rVert_{\text{col}}$,
normalizing each of the $D=1024$ columns across $N=784$ patches, following the official
\texttt{dense\_sparse\_matching.ipynb} protocol~\cite{simeoni2025dinov3}.

\paragraph{Unconstrained variant.}
For each foreground patch $p_i$ in face $A$:
$\hat{p} = \arg\max_j \tilde{f}_i \cdot \tilde{f}_j$
over all 784 patches of face $B$.
Semantic accuracy = fraction of matches in the same FaRL region as $p_i$.

\paragraph{FaRL-constrained variant.}
Search restricted to same-region patches of face $B$.
Spatial precision = fraction of matches within the median within-region pairwise
patch-grid distance (random baseline = 0.50).
All metrics bootstrapped over 200 pairs (1000 resamples, 95\% CIs).

\subsection{Temporal Label Propagation}

We implement the official \texttt{propagate()} protocol~\cite{simeoni2025dinov3}:
per-patch L2 normalization; anchor frame 0 (FaRL-labeled) always in context plus rolling
queue of last $Q=7$ frames; circular neighbourhood mask radius $r=12$; top-$K=5$ matches;
softmax temperature $\tau=0.2$; soft label probability propagation.
Metric: mean per-frame accuracy of predicted vs.\ ground-truth FaRL label over
foreground patches, averaged over frames 1--30 and over all 200 videos.
\textbf{No-FaRL ablation:} anchor initialized with uniform $q_j=1/M$ ($M=8$),
collapsing argmax to class 0 for all patches.

\subsection{Setup}

All experiments use CelebDF-v2 real videos~\cite{li2020celebdf} (YouTube-real split).
Region confusion and cross-identity matching use static frame pairs (segment 0, frame 0) from distinct identities.
Temporal tracking uses 200 videos, 30 consecutive frames each.
DINOv3 ViT-L/16: $448\times448$ input, 784 patches, $D=1024$; block-18 with BN
normalization, block-24 with LN-affine.
FaRL \texttt{lapa/448}: 8 foreground regions.
Normalization applied at query time only, never cached.
All experiments run on a Mac Pro 2019 with AMD Radeon Pro discrete GPU (MPS via Metal, not Apple Silicon); FaRL on CPU.

\section{Region-Level Correspondence Emerges Across Identity}

\subsection{Region Confusion Matrix}
\label{sec:c1}

\begin{figure}[t]
  \centering
  \includegraphics[width=\linewidth]{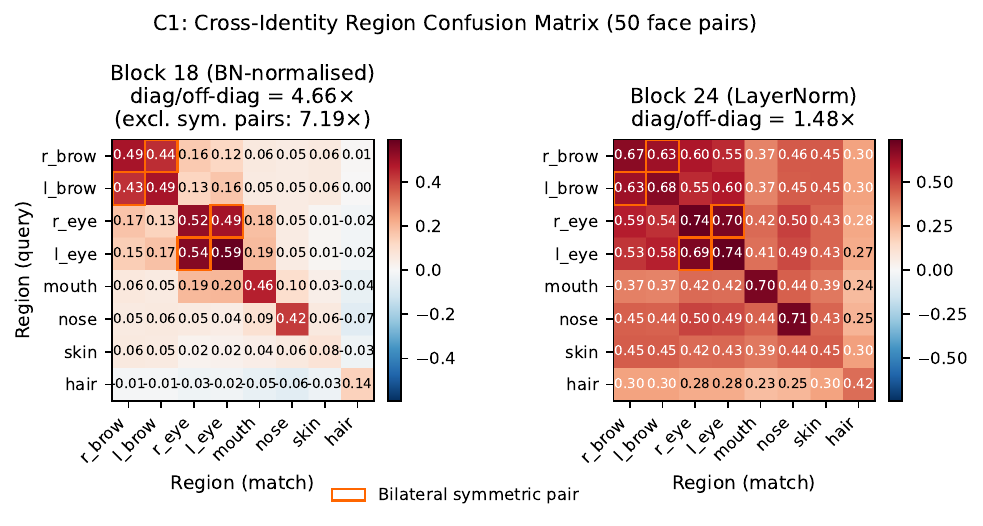}
  \caption{$8\times8$ cross-identity region confusion matrix (50 face pairs).
  Block-18 diagonal is $4.93\times$ the off-diagonal (rising to $7.19\times$ excluding
  the symmetric pairs in orange boxes). Block-24 collapses to $1.48\times$ due to
  global LayerNorm mixing.}
  \label{fig:c1}
\end{figure}

Results are shown in Figure~\ref{fig:c1} and Table~\ref{tab:c1}.
Block-18 achieves a $4.93\times$ ratio.
The elevated off-diagonal entries for r\_brow$\leftrightarrow$l\_brow (0.436) and
r\_eye$\leftrightarrow$l\_eye (0.492) reflect bilateral symmetry, not feature
noise; excluding these yields $7.19\times$.
Block 24 shows only $1.48\times$: the final token-wise LayerNorm projects all tokens
toward the unit hypersphere and mixes token directions, blending the per-dimension
profiles that keep regions directionally distinct into a global face representation.

\begin{table}[h]
  \caption{Region confusion ratios. Ratio = diag mean / off-diag mean.
  Symmetric-pair exclusion columns are computed on an independent 50-pair sampling
  (\texttt{c1\_symmetric\_analysis}; b18 diag 0.398), whose all-pairs ratio
  ($4.7\times$) is consistent with the main sampling.}
  \label{tab:c1}
  \centering
  \begin{tabular}{lccccc}
    \toprule
    Block & Diag & Off-diag & Ratio & Off-diag (excl.\ sym.) & Ratio (excl.\ sym.) \\
    \midrule
    b18 (BN) & 0.414 & 0.084 & \textbf{4.93$\times$} & 0.055 & \textbf{7.19$\times$} \\
    b24 (LN) & 0.637 & 0.429 & 1.48$\times$ & 0.411 & 1.55$\times$ \\
    \bottomrule
  \end{tabular}
\end{table}

\subsection{Cross-Identity Semantic Accuracy}

\begin{figure}[t]
  \centering
  \includegraphics[width=0.88\linewidth]{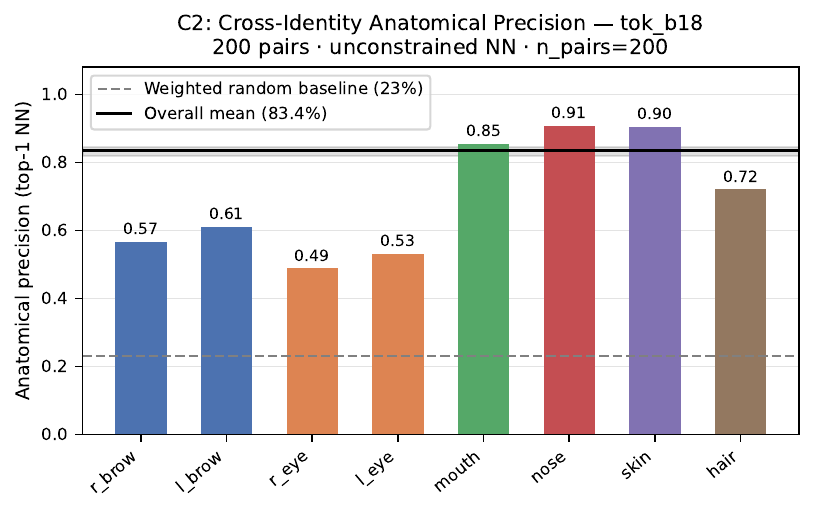}
  \caption{Per-region semantic accuracy for tok\_b18 (200 pairs, unconstrained NN).
  Dashed: weighted random baseline (23.0\%). Solid: overall mean (83.4\%).
  Brow and eye regions achieve 0.49--0.61 against $\leq$0.4\% random chance.}
  \label{fig:c2}
\end{figure}

Results are shown in Figure~\ref{fig:c2} and Table~\ref{tab:c2}.
Overall semantic accuracy: b18 $0.830$, b24 $0.823$, vs.\ weighted random $0.230$.
The random baseline is dominated by skin (57\% of foreground patches, random$=0.29$) and
hair (35\%, random$=0.18$); small regions (eyes, brows) have random baselines
$\leq 0.4\%$.
Nose (0.974 b24) and mouth (0.919 b24) are the most reliably matched mid-face regions.
The b18/b24 crossover is region-dependent: b18 outperforms b24 on brows (local structure
favors intermediate layers); b24 outperforms on mouth and nose (global context aids
structurally stable regions).
b18 and b24 are statistically tied on spatial precision (FaRL-constrained: 0.801 vs.\
0.805).
Qualitative examples of the matching, temporal-propagation, and clustering protocols
are shown in Appendix~\ref{app:qual}.

\begin{table}[h]
  \caption{Semantic accuracy (unconstrained, random=0.230) and spatial precision
  (FaRL-constrained, random=0.50) for 200 cross-identity pairs.}
  \label{tab:c2}
  \centering
  \small
  \begin{tabular}{lcccccc}
    \toprule
    & \multicolumn{3}{c}{Unconstrained} & \multicolumn{2}{c}{FaRL-constrained} \\
    \cmidrule(lr){2-4}\cmidrule(lr){5-6}
    Region & Random & b18 [95\% CI] & b24 [95\% CI] & b18 [95\% CI] & b24 [95\% CI] \\
    \midrule
    r\_brow & 0.004 & 0.597 [.548,.650] & 0.372 [.323,.424] & 0.279 [.228,.337] & 0.256 [.201,.311] \\
    l\_brow & 0.004 & 0.585 [.533,.637] & 0.437 [.384,.497] & 0.255 [.203,.308] & 0.257 [.205,.311] \\
    r\_eye  & 0.002 & 0.443 [.367,.518] & 0.544 [.468,.619] & 0.100 [.047,.167] & 0.078 [.028,.136] \\
    l\_eye  & 0.002 & 0.693 [.622,.762] & 0.523 [.446,.596] & 0.090 [.045,.146] & 0.062 [.028,.102] \\
    mouth   & 0.012 & 0.878 [.857,.897] & 0.919 [.894,.940] & 0.362 [.320,.402] & 0.342 [.301,.390] \\
    nose    & 0.019 & 0.920 [.910,.929] & 0.974 [.967,.980] & 0.449 [.399,.497] & 0.454 [.400,.503] \\
    skin    & 0.292 & 0.901 [.896,.905] & 0.874 [.868,.879] & 0.901 [.889,.914] & 0.912 [.894,.928] \\
    hair    & 0.183 & 0.723 [.683,.762] & 0.712 [.667,.756] & 0.768 [.729,.805] & 0.763 [.720,.804] \\
    \midrule
    \textbf{FG overall} & \textbf{0.230} &
      \textbf{0.830} [.814,.845] & 0.823 [.808,.838] &
      0.801 [.779,.822] & 0.805 [.781,.825] \\
    \bottomrule
  \end{tabular}
\end{table}

\paragraph{Cross-backbone comparison.}
Replicating cross-identity matching with frozen CLIP ViT-L/14 ($32{\times}32$ patch grid)
mirrors the temporal cross-backbone gap (Table~\ref{tab:c2crossvfm}).
Under the unconstrained protocol CLIP achieves only $0.678$ overall ($-15.2$~pp
vs.\ DINOv3-b18) and $0.433$ on the face-only mean ($-25.3$~pp), with the largest
deficits on l\_eye ($-45.9$~pp) and l\_brow ($-29.2$~pp).
Under the FaRL-constrained variant the face-only gap collapses to $\approx 0.4$~pp
($0.260$ for CLIP vs.\ $0.256$ for DINOv3-b18): once told ``this is a brow,''
CLIP can localize within the region. The asymmetry sharpens the interpretation:
CLIP encodes facial layout but cannot discriminate brow patches from non-brow
patches across identities. DINOv3 provides both anatomical identification and
spatial alignment; CLIP provides only the latter.
This implies the temporal face-only advantage (Table~\ref{tab:c3sweep}) is specifically
about anatomical identification, not spatial encoding.

\begin{table}[h]
  \caption{Cross-backbone semantic accuracy (200 cross-identity pairs). \emph{Face-only} = mean
  over r/l brows, r/l eyes, mouth, nose (skin/hair excluded). CLIP closes the gap
  on constrained face-only matching (knows layout) but fails on unconstrained
  matching (cannot identify anatomy across identities).}
  \label{tab:c2crossvfm}
  \centering
  \small
  \begin{tabular}{lcccc}
    \toprule
    & \multicolumn{2}{c}{Unconstrained} & \multicolumn{2}{c}{FaRL-constrained} \\
    \cmidrule(lr){2-3}\cmidrule(lr){4-5}
    Backbone & Overall & Face-only & Overall & Face-only \\
    \midrule
    \textbf{DINOv3-b18} & \textbf{0.830} & \textbf{0.686} & \textbf{0.801} & 0.256 \\
    DINOv3-b24          & 0.823 & 0.628 & 0.805 & 0.242 \\
    CLIP & 0.678 & 0.433 & 0.680 & \textbf{0.260} \\
    \midrule
    $\Delta$ (b18 $-$ CLIP) & $+15.2$ & $+25.3$ & $+12.1$ & $-0.4$ \\
    \bottomrule
  \end{tabular}
\end{table}

\section{The Correspondence Persists Through Time but Depends on the Right Semantics}

\subsection{Temporal Face-Part Tracking}

\begin{figure}[t]
  \centering
  \includegraphics[width=0.85\linewidth]{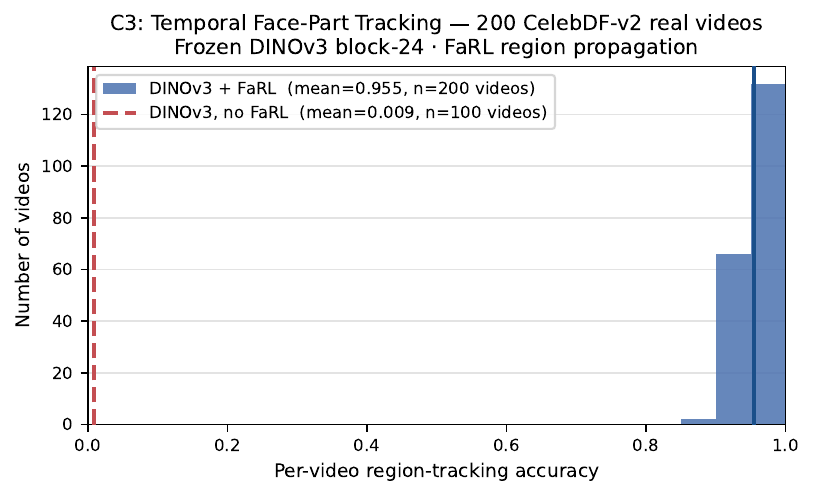}
  \caption{Per-video region-tracking accuracy distribution (200 CelebDF-v2 real videos,
  frozen DINOv3 block-24 + FaRL propagation). No-FaRL ablation (dashed red, mean=0.009)
  collapses entirely, isolating the two load-bearing components.}
  \label{fig:c3}
\end{figure}

Results are in Figure~\ref{fig:c3} and Table~\ref{tab:c3}.
Frozen DINOv3 + FaRL achieves $0.955 \pm 0.019$ across 200 videos.
Both blocks perform identically (0.955 vs.\ 0.954): temporal tracking depends on local
appearance consistency, encoded equally by both layers.
The no-FaRL ablation ($0.009 \pm 0.004$) is the most informative result: uniform anchor
initialization causes propagation to collapse to class 0 (r\_brow) for all patches,
revealing that FaRL initialization is entirely load-bearing for semantics and DINOv3 is
entirely load-bearing for spatial correspondence.
The PoC protocol (hard argmax, no neighbourhood, single-frame context) achieves only
$0.509$: the $+44.6$pp improvement from PoC to official protocol comes from the
measurement protocol, not the features.

\begin{table}[h]
  \caption{Temporal tracking results (200 CelebDF-v2 real videos).}
  \label{tab:c3}
  \centering
  \begin{tabular}{llcc}
    \toprule
    Protocol & Block & Mean & Std \\
    \midrule
    FaRL-initialized (official \texttt{propagate()}) & b24 & \textbf{0.955} & $\pm$0.019 \\
    FaRL-initialized (official \texttt{propagate()}) & b18 & 0.954 & $\pm$0.020 \\
    No-FaRL (uniform anchor)                         & b24 & 0.009 & $\pm$0.004 \\
    Hard argmax, no protocol (PoC baseline)          & b24 & 0.509 & $\pm$0.131 \\
    \bottomrule
  \end{tabular}
\end{table}

\paragraph{Block-depth and cross-model sweep.}
Table~\ref{tab:c3sweep} reports temporal tracking results across all five DINOv3 blocks and
CLIP ViT-L/14~\cite{radford2021clip} ($32{\times}32$ patch grid) under three metrics:
the patch-weighted aggregate (\emph{Agg.}), a \emph{balanced} metric (uniform mean over all 8 regions),
and a \emph{face-only} metric (uniform mean over r/l brows, r/l eyes, mouth, nose; skin and hair excluded).
Skin and hair together account for 91\% of foreground patches, structurally dominating the aggregate.

The aggregate is flat across all DINOv3 blocks ($0.953$--$0.955$, $\pm$0.2~pp), but face-only
reveals a monotonic decline: b18 peaks at $0.731$ and b24 reaches only $0.634$ ($-9.7$~pp),
exposing degradation of anatomical tracking quality through later layers that the aggregate conceals.
Against CLIP, DINOv3-b18 shows $+1.8$~pp aggregate advantage but $+16.8$~pp face-only
--- a ${\sim}9{\times}$ amplification from headline to geometrically-sensitive metric.
DINOv3's advantage concentrates in eyes ($+24$~pp) and brows ($+19$--$20$~pp);
both models track skin and hair comparably ($\leq$2~pp gap), confirming that CLIP's
image-level contrastive training is sufficient for coarse region tracking but insufficient
for the fine-grained anatomical correspondence that temporal facial analysis requires.

\begin{table}[h]
  \caption{Block-depth and cross-model sweep on temporal tracking (200 CelebDF-v2 real videos;
  FaRL-initialized \texttt{propagate()}).
  \emph{Balanced}: uniform mean over 8 regions.
  \emph{Face-only}: r/l brows, r/l eyes, mouth, nose (skin/hair excluded).}
  \label{tab:c3sweep}
  \centering
  \begin{tabular}{lccc}
    \toprule
    Backbone & Agg.\ mean & Balanced & Face-only \\
    \midrule
    DINOv3-b16 & 0.953 & 0.785 & 0.723 \\
    \textbf{DINOv3-b18} & \textbf{0.954} & \textbf{0.791} & \textbf{0.731} \\
    DINOv3-b20 & 0.955 & 0.773 & 0.706 \\
    DINOv3-b22 & 0.955 & 0.765 & 0.695 \\
    DINOv3-b24 & 0.955 & 0.721 & 0.634 \\
    \midrule
    CLIP & 0.936 & 0.662 & 0.563 \\
    \bottomrule
  \end{tabular}
\end{table}

\section{Controls and Mechanistic Checks}

\subsection{Attention Facet Ablation}

Following Amir et al.~\cite{amir2022deepvitfeaturesdense} (DINO keys $>$ tokens for
object correspondence), we hook \texttt{blocks[17].attn.qkv} and
\texttt{blocks[23].attn.qkv} to capture pre-RoPE projections and evaluate cross-identity
unconstrained precision (Table~\ref{tab:attn}).
Token outputs outperform all pre-RoPE projections: tok\_b18 achieves the highest overall
(0.834), reversing Amir et al.'s finding.
Among projections, values $>$ keys $>$ queries for both blocks.
DINOv3 applies RoPE to keys at query time; pre-RoPE keys are content-only and lack
positional phase required for spatial consistency.
The b18/b24 brow dissociation persists across all facet types, confirming a block-depth
effect rather than an aggregation artefact.

\begin{table}[h]
  \caption{Attention facet ablation on cross-identity matching (200 pairs, unconstrained). tok\_b18 achieves
  highest overall precision, reversing the DINO key $>$ token ordering.}
  \label{tab:attn}
  \centering
  \small
  \begin{tabular}{lcccccccc}
    \toprule
    Facet & Overall & r\_brow & l\_brow & r\_eye & l\_eye & mouth & nose & skin \\
    \midrule
    q\_b18 & 0.789 & 0.567 & 0.634 & 0.595 & 0.619 & 0.686 & 0.856 & 0.871 \\
    k\_b18 & 0.804 & 0.531 & 0.496 & 0.582 & 0.571 & 0.717 & 0.865 & 0.889 \\
    v\_b18 & 0.824 & 0.555 & 0.599 & 0.542 & 0.558 & 0.782 & 0.879 & 0.884 \\
    q\_b24 & 0.821 & 0.409 & 0.466 & 0.506 & 0.469 & 0.792 & 0.915 & 0.898 \\
    k\_b24 & 0.808 & 0.441 & 0.444 & 0.342 & 0.387 & 0.766 & 0.867 & 0.888 \\
    v\_b24 & 0.832 & 0.387 & 0.392 & 0.561 & 0.480 & 0.836 & 0.888 & 0.889 \\
    \textbf{tok\_b18} & \textbf{0.834} & \textbf{0.566} & \textbf{0.611} & 0.489 & 0.531 & 0.855 & 0.908 & \textbf{0.905} \\
    tok\_b24 & 0.826 & 0.372 & 0.408 & \textbf{0.618} & \textbf{0.513} & \textbf{0.917} & \textbf{0.970} & 0.879 \\
    \bottomrule
  \end{tabular}
\end{table}

\subsection{Independent Labeler Check}
\label{sec:independent}

To verify the cross-identity matching precision is not an artefact of FaRL serving as both method and ground
truth, we evaluate matching (b24, 200 pairs) under SegFormer-b5~\cite{xie2021segformer}
pretrained on CelebAMask-HQ (19 classes, human-annotated, independent of LaPa and FaRL).
FaRL--SegFormer inter-labeler agreement: $89.1\%$ on 100 frames.
Large and mid-face regions are stable: nose (0.974 vs.\ 0.980), skin (0.874 vs.\ 0.867),
mouth (0.919 vs.\ 0.873).
Eye and brow divergence at the 1--2 patch scale reflects label boundary differences
between CelebAMask-HQ and LaPa, not feature failure.

\subsection{Unsupervised Validation}
\label{sec:unsup}

To verify that FaRL labels reflect genuine structure in the DINOv3 feature space
rather than an evaluation artifact, we apply an Amir et al.~\cite{amir2022deepvitfeaturesdense}-style
protocol: pool all foreground DINOv3 patch descriptors from the 200 evaluation frames,
run $k$-means ($k \in \{4, 8, 12\}$, no FaRL supervision), and measure NMI and ARI of
the discovered clusters against FaRL region labels as independent ground truth
(Table~\ref{tab:unsup}).
Block-18 achieves NMI\,$=0.45$ ($k=8$), stable across $k \in \{4, 8, 12\}$ (range
$0.42$--$0.52$); cluster purity of $0.83$ indicates each discovered cluster is dominated
by a single FaRL region.
At $k=4$, NMI rises to $0.52$, consistent with DINOv3 encoding four macro-groups
(upper face, nose, mouth, surface) that FaRL's 8-region scheme subdivides.
Block-24 shows comparable clustering alignment (NMI\,$=0.45$, purity\,$=0.88$).
These results establish that FaRL labels are not an arbitrary evaluation choice:
DINOv3 independently organises facial patches into the same semantic groups, and FaRL
provides a named interface for structure the model encodes without supervision.
The clustering equivalence of b18 and b24 does not contradict the $4.93\times$
vs.\ $1.48\times$ finding: the block-18 advantage is an amplitude discriminability property;
$k$-means on L2-normalised features probes spatial cluster structure where amplitude is removed.

\begin{table}[h]
  \caption{Unsupervised validation: $k$-means (no FaRL) vs.\ FaRL labels as ground truth.
  200 frames, 80{,}973 foreground patches.}
  \label{tab:unsup}
  \centering
  \begin{tabular}{lcccccc}
    \toprule
    & \multicolumn{2}{c}{$k=4$} & \multicolumn{3}{c}{$k=8$} & $k=12$ \\
    \cmidrule(lr){2-3}\cmidrule(lr){4-6}\cmidrule(lr){7-7}
    Block & NMI & ARI & NMI & ARI & Purity & NMI \\
    \midrule
    b18 (BN) & 0.524 & 0.429 & 0.450 & 0.230 & 0.830 & 0.422 \\
    b24 (LN) & 0.482 & 0.386 & 0.453 & 0.223 & 0.879 & 0.415 \\
    Random   & $\approx$0 & $\approx$0 & $\approx$0 & $\approx$0 & 0.13 & $\approx$0 \\
    \bottomrule
  \end{tabular}
\end{table}

\section{Related Work}

\paragraph{Frozen ViT features for dense correspondence.}
Amir et al.~\cite{amir2022deepvitfeaturesdense} demonstrated zero-shot part-level
correspondence using frozen DINO-ViT keys (56.48\% PCK on
SPair-71k~\cite{zhang2025semanticcorrespondenceunifiedbenchmarking}).
PCK is the field-standard metric for semantic correspondence; it measures geometric
proximity to annotated keypoints and requires human-curated landmark pairs.
All prior frozen-feature correspondence work operates at the keypoint level.
We are the first to evaluate frozen VFM features at the face-part region level,
introducing region-level semantic accuracy as a direct test of whether the feature
space respects semantic region identity --- rather than geometric proximity to
annotated landmarks.

\paragraph{Self-supervised temporal correspondence.}
Wang et al.~\cite{wang2019learningcorrespondencecycleconsistencytime} showed
cycle-consistency in time provides free supervisory signal for correspondence.
Jabri et al.~\cite{jabri2020spacetimecorrespondencecontrastiverandom} formalized this
via contrastive random walk, learning features that transfer to video object segmentation.
DINOv3's Gram-anchored training is a stronger instance; our temporal tracking result ($95.5\%$) shows
it is sufficient for face tracking without temporal training.

\paragraph{Diffusion features for correspondence.}
Tang et al.~\cite{tang2023emergentcorrespondenceimagediffusion} showed Stable Diffusion
U-Net features support semantic correspondence without fine-tuning.
Zhang et al.~\cite{zhang2023talefeaturesstablediffusion} showed SD and DINO features
are complementary for zero-shot semantic correspondence.
These confirm correspondence as a general emergent property of large-scale self-supervised
models~\cite{zhang2025semanticcorrespondenceunifiedbenchmarking}; our work makes the
mechanistic claim specific to DINOv3's Gram anchoring and establishes it at the
face-part region level for the first time.

\paragraph{Point tracking and trained upper bounds.}
TAP-Vid~\cite{doersch2023tapvidbenchmarktrackingpoint} introduced the point-tracking
benchmark and \texttt{propagate()} API we adopt for temporal tracking.
Chrono~\cite{kim2025chrono} augments frozen DINOv2 with a learned temporal adapter,
achieving state-of-the-art on TAP-Vid-DAVIS.
Our evaluation uses frozen DINOv3 with no temporal adapter evaluated on face-part
labels; the $95.5\%$ accuracy shows that face-specific Gram anchoring reduces the need
for temporal training in the face domain.

\section{Limitations and Conclusion}

Frozen DINOv3 ViT-L/16 features support region-level facial correspondence across
identities and time without any face-specific training. The strongest signal comes
from an intermediate layer: block 18 yields $4.93\times$ same-region versus
cross-region discrimination, $83.0\%$ unconstrained cross-identity matching
accuracy, and $95.5\%$ temporal tracking, while the final block is markedly more
mixed. The cross-backbone comparison clarifies what kind of structure is being
measured: CLIP closes the gap once the search is constrained to the correct
region, but lags by $+16.8$~pp on unconstrained anatomical matching. The
comparison is diagnostic rather than merely competitive --- DINOv3 provides
anatomical identification, not only spatial alignment.

Three limitations bound the claims. FaRL is both labeling interface and ground
truth, though a SegFormer replication (\S\ref{sec:independent}) and unsupervised
clustering~\cite{amir2022deepvitfeaturesdense} (\S\ref{sec:unsup}, NMI\,$=0.45$)
indicate the evaluation reflects feature-space structure rather than labeler
artefact. The evaluation uses a single dataset (CelebDF-v2) and parsing scheme
(LaPa). And the cross-identity matching test measures region-level semantic accuracy by design rather than
keypoint PCK, since cross-identity landmark pairs are not available for our
setting; \cite{zhang2025semanticcorrespondenceunifiedbenchmarking} identify
fine-tuning as the dominant performance factor in semantic correspondence, so
our frozen numbers should be read as a principled lower bound for the face
domain. Useful next steps include a DINOv3-native LaPa face parser, a mutual
best-buddy variant for cross-identity matching, and face-specific fine-tuning to establish the
supervised upper bound.

Conceptually, the contribution is to identify a frozen, intermediate-layer
coordinate system for facial regions: a representation in which anatomical
identity is preserved across people and through time, recoverable by simple
nearest-neighbor matching, without any face-specific supervision.

\section{Broader Impact}

Region-level facial correspondence in frozen vision foundation models supports
several beneficial applications: face normalization and recognition under pose
and expression variation, biometric and forensic identification, real-time
avatar animation in XR/AR/VR systems, and clinical analyses including facial
asymmetry measurement, post-surgical change tracking, syndrome diagnosis
(e.g., Noonan syndrome), and morphometric studies. Segmentation tells you
\emph{what} each pixel is; correspondence tells you \emph{where} it maps in
another face --- modern pipelines combine both. We acknowledge dual-use risk
(face manipulation, surveillance, unauthorized biometric tracking) and mitigate
it by introducing no novel correspondence algorithm and using only publicly
released frozen pre-trained models, releasing no new face data or weights.


\newpage
\appendix
\section{Qualitative Examples}
\label{app:qual}

\begin{figure}[htbp]
  \centering
  \includegraphics[width=\linewidth]{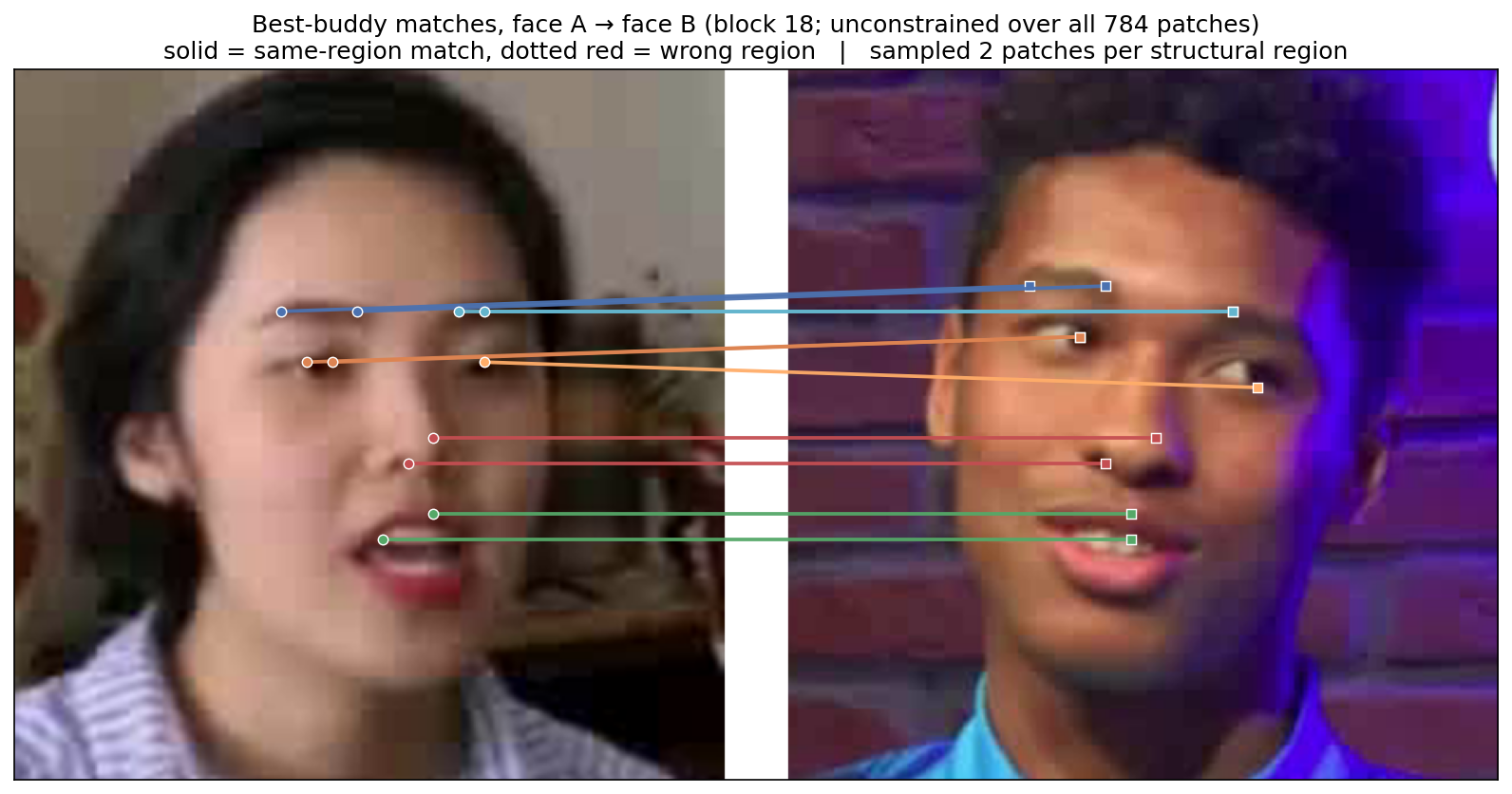}
  \caption{Unconstrained best-buddy matching between two identities (block 18,
  per-channel L2). Each sampled patch of face A (two per structural region) is
  connected to its nearest neighbor among all 784 patches of face B; the search
  is never told the target region. Solid lines: same-region matches; dotted red:
  wrong-region matches. Matches land on the correct anatomy despite the pose
  difference.}
  \label{fig:app-matchlines}
\end{figure}

\begin{figure}[htbp]
  \centering
  \includegraphics[width=\linewidth]{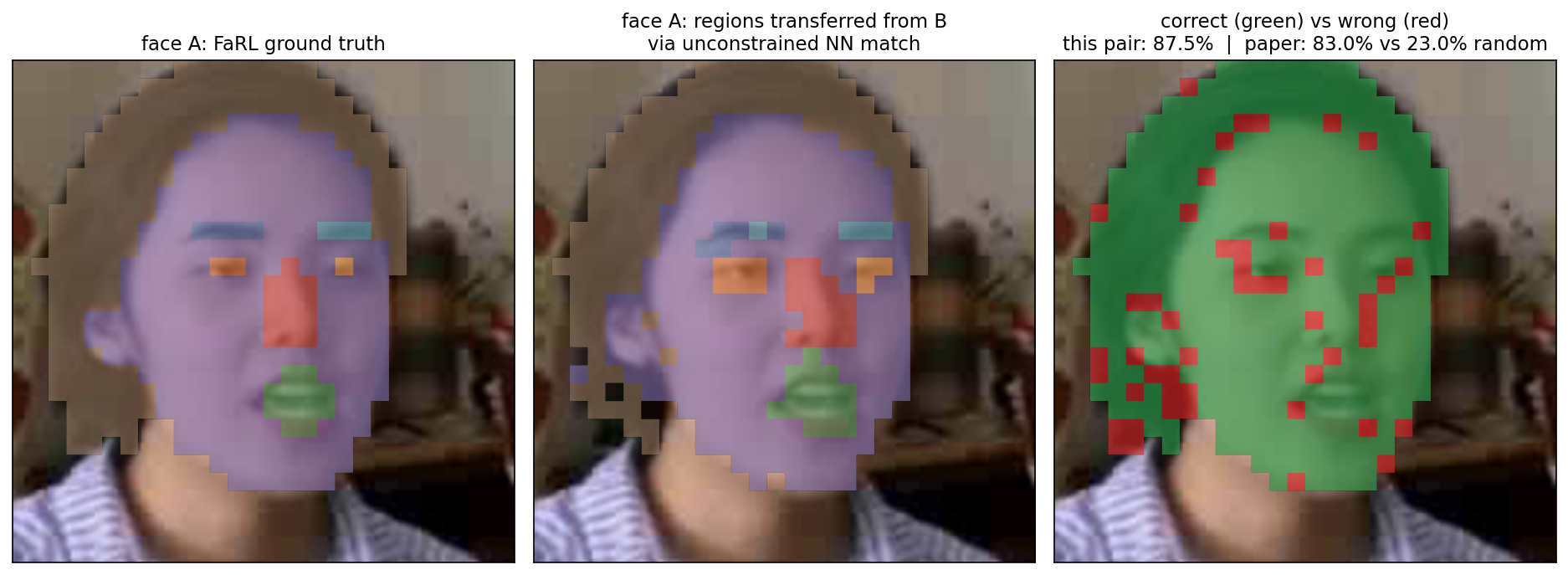}
  \caption{Dense label transfer for the same pair: every foreground patch of
  face A is recolored by the FaRL region of its best buddy in face B. Left: FaRL
  ground truth; middle: transferred regions; right: agreement map (green correct,
  red wrong). Accuracy on this pair is $87.5\%$; the 200-pair mean is $0.830$
  against the $0.230$ weighted random baseline (Table~\ref{tab:c2}).}
  \label{fig:app-transfer}
\end{figure}

\begin{figure}[htbp]
  \centering
  \includegraphics[width=0.92\linewidth]{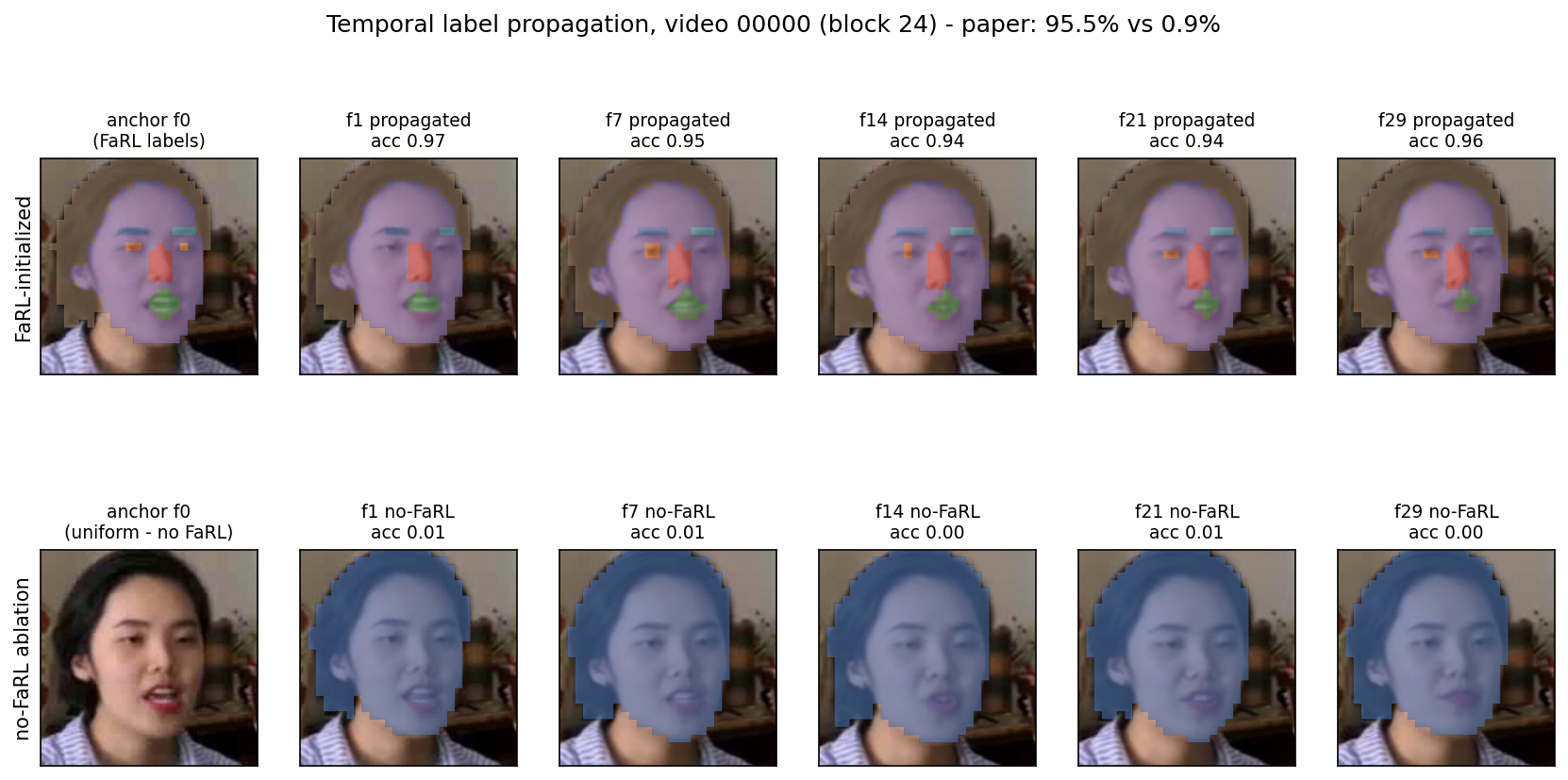}
  \caption{Temporal label propagation (block 24, official \texttt{propagate()}
  protocol). Top row: FaRL labels the anchor frame only; propagated labels track
  the face through frame 29 (this video: $0.951$; 200-video mean: $0.955$,
  Table~\ref{tab:c3}). Bottom row: the no-FaRL ablation replaces the anchor with a
  uniform distribution, and the same features and protocol collapse to a single
  class (this video: $0.008$; mean: $0.009$). FaRL is load-bearing for semantics;
  DINOv3 is load-bearing for spatial correspondence.}
  \label{fig:app-temporal}
\end{figure}

\begin{figure}[htbp]
  \centering
  \includegraphics[width=0.55\linewidth]{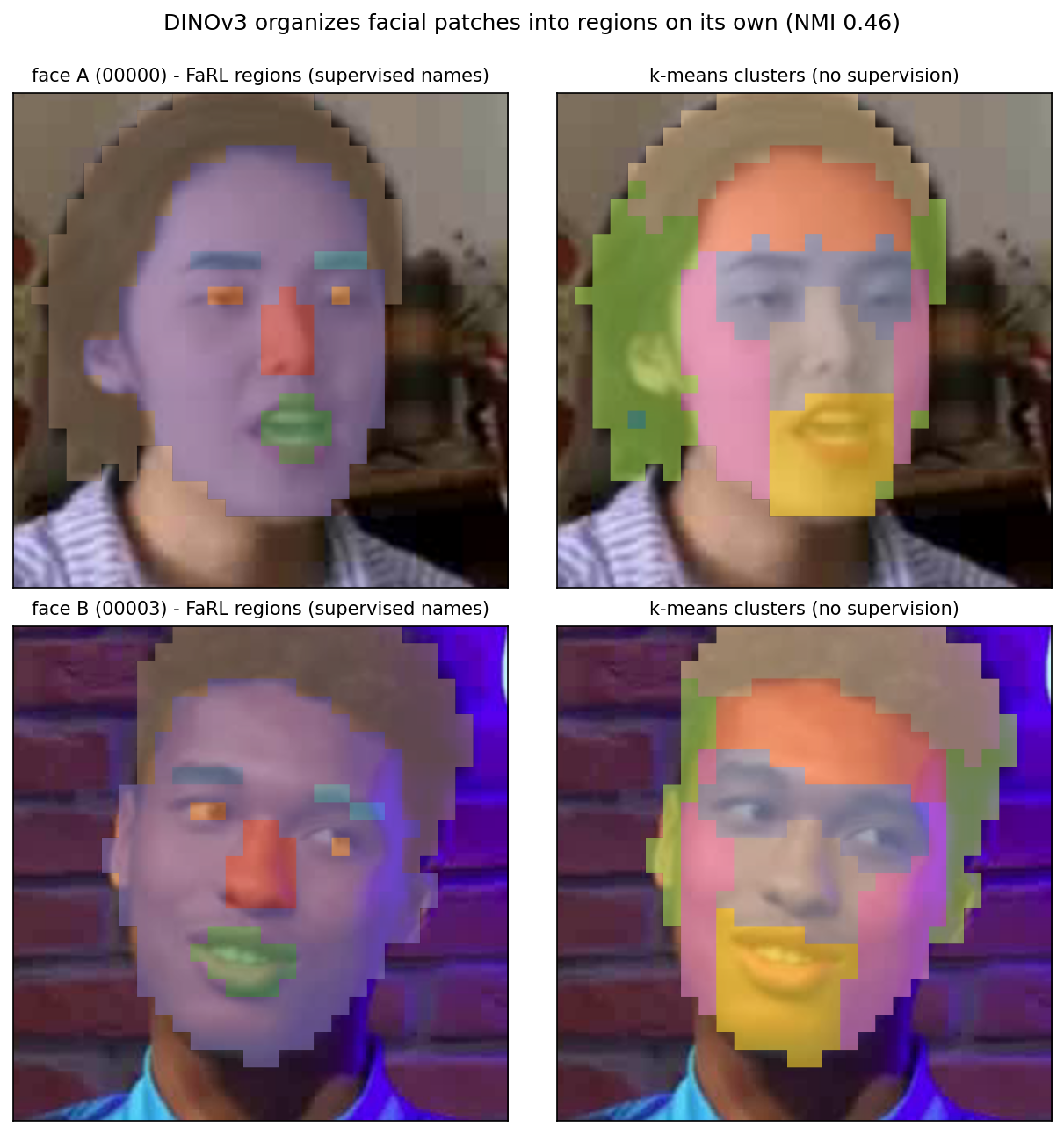}
  \caption{Unsupervised validation (\S\ref{sec:unsup}): $k$-means ($k=8$) on frozen
  block-18 patch features, fit without FaRL supervision, painted on the two demo
  identities next to the FaRL regions. Cluster identities are arbitrary
  (unsupervised palette); their spatial support recovers the facial regions
  (NMI $=0.459$ on this fit; $0.450$ in Table~\ref{tab:unsup}).}
  \label{fig:app-kmeans}
\end{figure}

This appendix illustrates the protocols of \S2 on individual samples from the
evaluation set (two identities from the CelebDF-v2 YouTube-real split). All panels
are rendered from the same cached features and FaRL patch labels used in the
quantitative experiments; per-panel numbers are single-pair (or single-video)
values, shown next to the corresponding evaluation-set statistics from the main
text. Figure~\ref{fig:app-matchlines} shows unconstrained cross-identity
best-buddy matching (the C2 protocol); Figure~\ref{fig:app-transfer} the resulting
dense region-label transfer; Figure~\ref{fig:app-temporal} temporal label
propagation together with its no-FaRL collapse (the C3 protocol); and
Figure~\ref{fig:app-kmeans} the unsupervised $k$-means validation of
\S\ref{sec:unsup}.

\end{document}